\documentclass[sigconf]{acmart}

\usepackage{bm}
\usepackage{multirow} 
\usepackage{threeparttable} 
\usepackage{xspace}
\usepackage{libertine}
\usepackage{tipa} 
\usepackage{balance}
\usepackage{libertine}
\usepackage[T3,T1]{fontenc}
\DeclareFontFamilySubstitution{T3}{LinuxLibertineT-TLF}{cmr}

\newcommand{\onedot}{%
  \if\relax\detokenize{.}\relax
  \else
    .\null%
  \fi
  \xspace
}

\newcommand{\ie}{\emph{i.e}\onedot}

\setcopyright{none}
\renewcommand\footnotetextcopyrightpermission[1]{}
\AtBeginDocument{%
  }

\setcopyright{acmlicensed}

\copyrightyear{2025}
\acmYear{2025}
\setcopyright{acmlicensed}\acmConference[MM '25]{Proceedings of the 33rd ACM International Conference on Multimedia}{October 27--31, 2025}{Dublin, Ireland}
\acmBooktitle{Proceedings of the 33rd ACM International Conference on Multimedia (MM '25), October 27--31, 2025, Dublin, Ireland}
\acmDOI{10.1145/3746027.3755568}
\acmISBN{979-8-4007-2035-2/2025/10}

\begin{document}

\title{PTalker: Personalized Speech-Driven 3D Talking Head Animation via Style Disentanglement and Modality Alignment}

\author{Bin Wang}
\affiliation{%
  \institution{College of Computer Science and Electronic Engineering \\ Hunan University}
  \city{Changsha}
  \country{China}
}
\email{bynnwang@hnu.edu.cn}

\author{Yang Xu}
\affiliation{%
  \institution{College of Computer Science and Electronic Engineering \\ Hunan University}
  \city{Changsha}
  \country{China}
}
\email{xuyangcs@hnu.edu.cn}

\author{Huan Zhao}
\affiliation{%
  \institution{College of Computer Science and Electronic Engineering \\ Hunan University}
  \city{Changsha}
  \country{China}
}
\email{hzhao@hnu.edu.cn}

\author{Hao Zhang}
\affiliation{%
  \institution{School of Electronic Information \\ Central South University}
  \city{Changsha}
  \country{China}
}
\email{209122@csu.edu.cn}

\author{Zixing Zhang}
\authornote{Corresponding author.}

\affiliation{%
  \institution{College of Computer Science and Electronic Engineering \\ Hunan University}
  \city{Changsha}
  \country{China}
}
\email{zixingzhang@hnu.edu.cn}

\renewcommand{\shortauthors}{Bin Wang et al.}

\begin{abstract}
Speech-driven 3D talking head generation aims to produce lifelike facial animations precisely synchronized with speech. While considerable progress has been made in achieving high lip-synchronization accuracy, existing methods largely overlook the intricate nuances of individual speaking styles, which limits personalization and realism. In this work, we present a novel framework for \textbf{personalized} 3D talking head animation, namely \textbf{``PTalker''}. This framework preserves speaking style through style disentanglement from audio and facial motion sequences and enhances lip-synchronization accuracy through a three-level alignment mechanism between audio and mesh modalities. Specifically, to effectively disentangle style and content, we design disentanglement constraints that encode driven audio and motion sequences into distinct style and content spaces to enhance speaking style representation. To improve lip-synchronization accuracy, we adopt a modality alignment mechanism incorporating three aspects: spatial alignment using Graph Attention Networks to capture vertex connectivity in the 3D mesh structure, temporal alignment using cross-attention to capture and synchronize temporal dependencies, and feature alignment by top-$k$ bidirectional contrastive losses and KL divergence constraints to ensure consistency between speech and mesh modalities. Extensive qualitative and quantitative experiments on public datasets demonstrate that PTalker effectively generates realistic, stylized 3D talking heads that accurately match identity-specific speaking styles, outperforming state-of-the-art methods. The source code and supplementary videos are available at: \href{https://acmmm25p.bwbwiwn.site/}{PTalker}.

\end{abstract}

\ccsdesc[500]{Computing methodologies~Computer vision tasks}
\ccsdesc[500]{Computing methodologies~Animation}

\keywords{Personalized Talking Head, 3D Facial Animation, Style Disentanglement, Modality Alignment}

\maketitle

\begin{figure}[!t]
  \centering
   \includegraphics[scale=0.45]{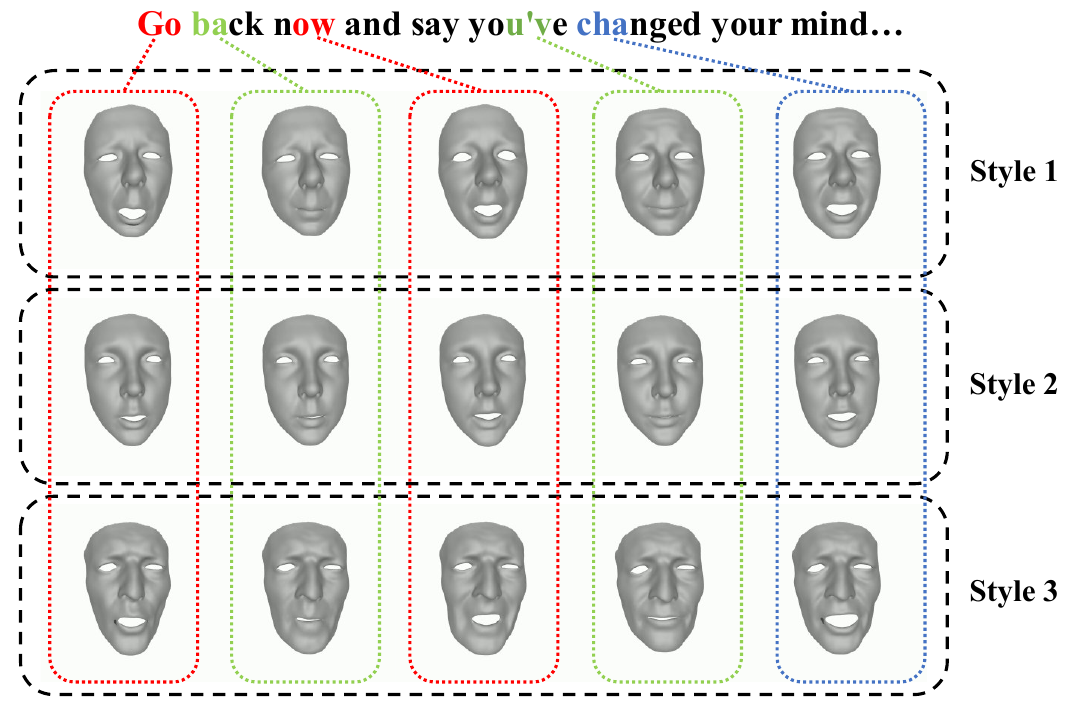}
   \caption{Demonstrations of diverse speaking styles, highlighting variations in mouth opening and closing amplitudes (outlined in red and green borders, respectively), and lip shape changes and plumpness, such as lip pouting or curling (outlined in blue border).}
   \label{fig:ss}
   \Description{.}
\end{figure}

\section{Introduction}
\label{sec:intro}

Speech-driven 3D talking heads are in high demand due to their wide range of applications, such as 3D avatars, virtual reality, and live streaming~\cite{pan2024expressive,wang2024trust,christoff2023application}. Recent advancements in deep learning have resulted in remarkable improvements in generating highly accurate lip-synchronized 3D talking heads. However, existing methods~\cite{richard2021meshtalk,fan2022faceformer,xing2023codetalker} have primarily focused on synchronizing 3D talking heads with the speech content, often overlooking individual speaking styles. This oversight limits the personalization and realism of the generated talking heads, which are crucial for user engagement, ultimately reducing the overall immersive experience.

Speaking style captures the unique facial motion patterns of a speaker. These patterns are characterized by parameters, such as the amplitude of mouth opening and closing, the degree of lip protrusion, and other related dynamics, as shown in Fig.~\ref{fig:ss}, critical for achieving realistic and personalized animations. Recent studies~\cite{cudeiro2019capture, song2024talkingstyle, fu2024mimic, zhou2024meta, yang2025stylespeaker} have increasingly highlighted the importance of speaking style in 3D talking head generation, achieving progress in creating personalized animations.

Previous work on speaking style can be categorized into two main approaches: the first involves training person-specific models for individual subjects, while the second maps speaking style to identity categories. The former approaches~\cite{karras2017audio,richard2021audio} can produce facial animations tailored to specific individuals' speaking styles but struggle to generalize across diverse subjects.
The latter approaches~\cite{cudeiro2019capture,richard2021meshtalk,fan2022faceformer,xing2023codetalker, wu2023speech, fu2024mimic} use one-hot encoding to differentiate speaking styles, enabling limited control over style representation. However, these approaches are still problematic: the former is constrained by the availability of subject-specific data and scope, while the latter's one-hot encoding fails to capture the complexity of style information, thereby lacking sufficient generalization to unseen subjects. Recent studies have focused on improving the style generalization ability to unseen speakers. For instance, Imitator~\cite{thambiraja2023imitator} improves its adaptability to unseen speakers by fine-tuning a generalized model, which results in additional overhead. TalkingStyle~\cite{song2024talkingstyle} improves one-hot encoding by modeling global and averaged styles but performs poorly when handling subjects with styles that deviate from training data. Some methods~\cite{wu2023speech, fu2024mimic} extract styles from facial motions but ignore speech context, leading to less accurate extraction and reduced adaptability to unseen speakers.

For this reason, this paper presents a novel framework, named ``PTalker'', for 3D talking head generation, which simultaneously enhances speaking style representation and improves lip-sync accuracy. For \textbf{speaking style}, rather than relying solely on one-hot encoding, we propose a disentanglement strategy that extracts style information from both speech and facial motion sequences. In this manner, comprehensive style codes are generated by fusing the one-hot encoding with disentangled style features.  This design should theoretically result in a richer style representation.

For \textbf{lip-sync}, we focus on refining the alignment between the 3D mesh and audio modalities to improve lip-sync accuracy. We introduce a three-level alignment mechanism that jointly addresses spatial, temporal, and feature-level mismatches between audio and 3D facial motion. In particular, Graph Attention Networks (GATs) are utilized for spatial alignment of the 3D mesh, cross-attention mechanisms are adopted for temporal synchronization, and a contrastive loss is applied to align features at the latent level.

Additionally, we incorporate a series of carefully designed constraints to achieve both effective disentanglement and cross-modal alignment. Specifically, the orthogonality loss and the mutual information minimization loss contribute to decouple style from content by discouraging redundant information sharing between the two representations. In parallel, the top-\(k\) contrastive loss enforces local alignment by ensuring that corresponding audio and motion content features are similar, while Kullback-Leibler (KL) divergence loss aligns their global statistical properties to ensure consistency of the overall content distributions. In addition, we integrate a style similarity loss and an adversarial loss to further refine the separation of style from content. Together, these constraints robustly disentangle style and content at both local and global levels, and ensure effective alignment between audio and motion modalities.

By integrating these techniques, PTalker can generate realistic 3D talking heads that are both lip-synchronized and faithfully reflect an individual’s speaking style. In summary, the main contributions are as follows:

$\bullet$ We propose PTalker, a novel framework capable of generating realistic 3D talking heads that align with identity-specific speaking styles from driven speech.

$\bullet$ We incorporate well-designed constraints to disentangle audio and motion into style codes and content space, combining with one-hot encoding to enhance speaking style representation.

$\bullet$ We introduce a three-level alignment mechanism that aligns audio and mesh modalities across spatial, temporal, and feature domains, reducing misalignment and improving lip-sync accuracy. 

$\bullet$ We conduct extensive qualitative and quantitative experiments on two widely used datasets, demonstrating that our method outperforms current state-of-the-art methods.

\section{Related Work}
\label{sec:related}

\begin{figure*}[!ht]
  \centering
   \includegraphics[scale=0.5]{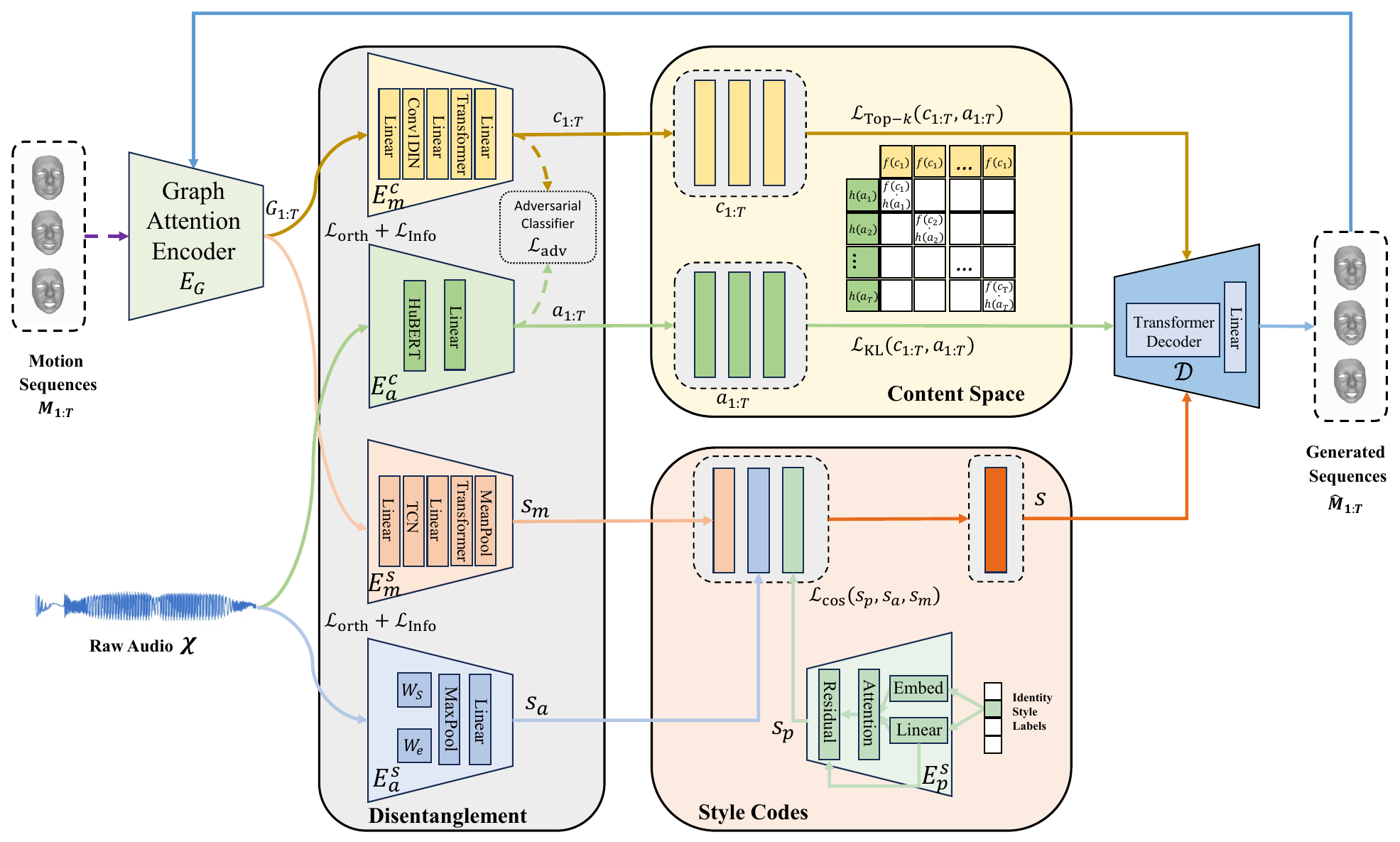}
   \caption{Illustration of PTalker. The input motion sequence $M_{1:T}$ is first encoded by the graph attention encoder $ E_G$ to produce $G_{1:T}$. Then, the motion style encoder $ E_{m}^{s}$ and the motion content encoder $E_{m}^{c}$ extract the motion style $s_m$ and content features $c_{1:T}$ from $G_{1:T}$. The synchronized raw speech \textbf{$\chi$} is encoded into audio features $a_{1:T}$ and audio style $s_a$ by the audio content encoder $E_{a}^{c}$ and the audio style encoder $E_{a}^{s}$. The identity style encoder $E_{p}^{s}$ encodes the one-hot vector into the personal style $s_p$, where $s_p$, combined with $s_m$ and $s_a$, produces the final style code $s$. The motion decoder $\mathcal{D}$ reconstructs the facial motions $\hat{M}_{1:T}$ by combining $s$ with $c_{1:T}$ and $a_{1:T}$. During inference, $E_{m}^{c}$ and $E_{m}^{s}$ encode the generated frame sequence from the driving audio to obtain the motion style and content. All dashed and solid lines in the figure together constitute the training process, while excluding the dashed lines represents the inference process.
   }
   \label{fig:framework}
   \Description{.}
\end{figure*}

\textbf{Speech-Driven 3D Talking Head Generation.} Speech-driven 3D talking head generation aims to synchronize facial dynamics with arbitrary speech to create vivid head animations, an area where significant progress has recently been made. Prior research typically falls into two categories: parameter-based and vertex-based approaches. Parameter-based methods~\cite{zhou2018visemenet,huang2021speaker,bao2023learning} establish complex mapping rules, often requiring extensive processing. In contrast, our approach focuses on vertex-based methods, which directly map speech to 3D mesh vertices. This enables more efficient capture of subtle facial expressions and yields superior results in animation generation.

VOCA~\cite{cudeiro2019capture} first introduced a CNN-based method that models multiple subjects with one-hot encoding to produce realistic talking heads, concentrating only on the mouth region. Meshtalk~\cite{richard2021meshtalk} expanded its focus to the entire face, capturing dynamics beyond just the lower face. FaceFormer~\cite{fan2022faceformer} introduced a Transformer architecture to predict displacements on a template mesh, ensuring temporal stability, while CodeTalker~\cite{xing2023codetalker} used discrete facial spaces to achieve a more precise speech-facial correlation. Building on FaceFormer, many optimized Transformer-based methods~\cite{su2023dualtalker,wu2024mmhead,jafari2024jambatalk} have emerged; however, they still encounter error accumulation due to auto-regression and suffer from over-smoothing. To address this, some methods have incorporated the diffusion mechanism to mesh vertices, showing promising results. FaceDiffuser~\cite{stan2023facediffuser} was the first end-to-end model to explicitly apply diffusion to mesh vertices, producing realistic and diverse animation sequences. DiffSpeaker~\cite{ma2024diffspeaker} combined Transformer and diffusion models to solve the audio-4D data scarcity issue, enabling real-time and efficient generation. 3DiFACE~\cite{thambiraja20253diface} proposed a fully-convolutional diffusion model and a novel sparsely-guided diffusion to enable precise control and editing for generated animation. Although these methods have achieved notable results, there is still room for improvement in lip-sync accuracy and personalization. To address these gaps, PTalker employs a style disentanglement and a three-level alignment for further enhancement.
  
\textbf{Speaking Style in 3D Talking Head Generation} Speaking style, which encompasses both individual style and emotional expression, can generally be divided into two categories: ``label-based styles'' and ``context-based styles''. Label-based styles use a fixed set of categories as labels, often one-hot encoding~\cite{fan2022faceformer, song2024talkingstyle, chen2025diffusiontalker, yang2025stylespeaker}. For example, TalkingStyle~\cite{song2024talkingstyle} combined the common and the personalized style to make the speaking style more distinct. DiffusionTalker~\cite{chen2025diffusiontalker} utilized the personalizer-guided distillation to enhance accurate lip language with personalized speaking styles. StyleSpeaker~\cite{yang2025stylespeaker} explicitly extracts speaking styles according to speaker characteristics while adjusting for style biases resulting from different speeches. Context-based methods, on the other hand, require reference snippets or specific actions from which new speaking styles can be inferred. Imitator~\cite{thambiraja2023imitator} and AdaMesh~\cite{chen2025adamesh} used 2D video references to learn style embeddings, while DiffPoseTalk~\cite{sun2024diffposetalk} extracted style from provided facial animations. MetaFace~\cite{zhou2024meta} achieved speaking style adaptation with minimal data through meta-learning. Moreover, disentanglement techniques are effective for separating style information. EmoTalk~\cite{peng2023emotalk} and EmoFace~\cite{lin2024emoface} explored isolating emotion style from speech to realize emotional speech-driven facial motion. CASTalker~\cite{liu2024content} decoupled content and style representations from audio and corresponding text to make it controllable. However, fully disentangling style and content is challenging. In this work, we aim to decouple and capture unique speaking styles of unseen subjects from driven speech by well-designed constraints and networks.

\section{Methodology}
\label{sec:meth}
\subsection{Network Structures}
We propose PTalker to achieve style-content disentanglement and audio-mesh modality alignment, enabling the synthesis of talking head animations with high lip-sync accuracy that reflect corresponding speaking styles, as illustrated in Fig.~\ref{fig:framework}. During training, we aim to disentangle style and content features from both the raw audio $\chi$ and the corresponding facial motion sequences $M_{1:T}$ with $T$ frames. The audio style and motion style are subsequently combined with the identity style to form the final style codes. Meanwhile, the content from both audio and motion is integrated into the content space. Additionally, we introduce a three-level alignment mechanism that addresses spatial (the graph attention encoder $E_G$), temporal (the motion decoder $\mathcal{D}$), and feature-level ($\mathcal{L}_{\text{cts}}$) discrepancies between the audio and 3D mesh domains. 

\textbf{Graph Attention Encoder.}
The feature representation of a mesh vertex depends on its own features as well as those of its neighboring vertices. We introduce the GAT encoder $E_{G}$ to effectively capture local structure information. GAT updates the vertex coordinate matrix $V \in \mathbb{R}^{N_v \times 3}$ by aggregating the features of adjacent vertices at each layer, where $N_v$ denotes the number of vertices. By stacking multiple layers of GAT, each vertex can integrate information from its neighborhood comprehensively, achieving spatial alignment. The output feature $G_{1:T} = E_{G} (M_{1:T})$ represents the updated feature of each vertex after integrating neighborhood context. This feature is then used for motion disentanglement, where $T$ denotes the number of frames.

\textbf{Audio Style-Content Disentanglement.} We introduce an \textbf{audio style encoder $E_{a}^{s}$} and an \textbf{audio content encoder $E_{a}^{c}$} to effectively disentangle the style and content features from raw audio $\chi$. Specifically, $E_{a}^{s}$ is initialized with two pre-trained Wav2Vec2~\cite{baevski2020wav2vec} sub-networks: $W_s$ for speaker identification and $W_e$ for emotion classification, enabling the extraction of audio style code $s_{a}$. The audio content encoder $E_{a}^{c}$ extracts content features $a_{1:T} \in \mathbb{R}^{T \times D_{a}}$ from $\chi$, initialized with pre-trained HuBERT~\cite{hsu2021hubert} weights. Disentanglement is achieved because $W_s$ and $W_e$ are trained on content-independent tasks, while $E_{a}^{c}$ leverages pre-training on diverse content datasets. Additionally, the audio orthogonal loss $\mathcal{L}_{\text{orth}}$ and the mutual information minimization loss $\mathcal{L}_{Info}$ strengthen this process by enforcing the separation of $s_{a}$ and $a_{1:T}$ distributions.

\textbf{Motion Style-Content Disentanglement.} 
The \textbf{motion style encoder} $E_{m}^{s}$ aims to obtain a compact latent style code $s_{m}\in\mathbb{R}^{D_{sm}}$ from $G_{1:T}$, which captures unique style characteristics from the input, invariant to temporal variations. Here, we first project $G_{1:T}$ to a lower-dimensional space via a linear mapping layer, compressing the feature representation to make it suitable for subsequent processing. Then, we refine these features using a multi-layer 1D temporal convolution network (TCN), which creates style tokens by capturing local temporal patterns within the motion sequence. Next, to model longer-range temporal dependencies and derive stylistic representations among these tokens, we employ a multi-layer Transformer encoder. Finally, we aggregate the style information using temporal mean pooling, producing a compact motion style $s_{m}$ effectively that encapsulates distinct style properties.

In contrast, the \textbf{motion content encoder} $E_{m}^{c}$ adopts a similar structure to $E_{m}^{s}$ but is specifically designed to capture dynamic, time-varying content features in the input motion sequence. Initially, the input $G_{1:T}$ is mapped to a lower-dimensional feature space via a fully connected layer. Then, it is processed through a series of instance-normalized convolutional layers (Conv1DIN) to extract local temporal patterns. Next, the encoded features are projected to a hidden space with positional encoding, followed by a transformer encoder generating content representations. Finally, we obtain the latent content features $c_{1:T} \in \mathbb{R}^{T \times D_{m}}$ through a linear projection. Additionally, we also apply $\mathcal{L}_{\text{orth}}$ and $\mathcal{L}_{\text{Info}}$ between $s_{m}$ and $c_{1:T}$ to enforce decoupling the motion style and content distributions.

\textbf{Identity Style Encoder.} In previous implementations, one-hot style labels are straightforwardly encoded, which binds the style containing limited information. Therefore, we introduce an \textbf{identity style encoder} to enhance the style representation. Firstly, we extract two style embeddings: the personalized style, which maps the input one-hot identity vector to a feature embedding using a linear layer, and the common style, which generates identity-specific embeddings based on the input's index. Then, we apply an attention mechanism to capture important identity features between the personalized and the common styles. Finally, we utilize a residual connection to combine the original personalized style with the attention output and get the final identity-specific style features $s_{p}$.
Therefore, we obtain the final multi-dimensional style codes $s = s_{a} + s_{m} + s_{p}$, which can better represent the style information.

\textbf{Motion Decoder.}
The motion decoder $\mathcal{D}$ produces sequential facial motions $\hat{M}_{1:T}$ by combining the style code $s$, the content features $c_{1:T}$ and the audio features $a_{1:T}$. Specifically, $\mathcal{D}$ is an auto-regressive model that generates motion sequences frame-by-frame using a transformer decoder. In each step, the model applies periodic positional encoding and a biased mask to control attention. Then, the transformer decoder mainly consists of causal multi-head self-attention and multi-head cross-attention, allowing it to iteratively build the motion sequence and align the audio-motion modalities.

\subsection{Disentanglement Constraints}

\textbf{Adversarial Classifier Loss}. To improve the disentanglement between content and style information, we introduce an adversarial loss that leverages a style classifier after content encoders to enforce the removal of style-related information from the content representations. Talking the output of the audio content encoder (\ie $a_{1:T}$ ) as an example, we first obtain a global content representation $a_{c}$ by temporal averaging on $a_{1:T}$. Then, we apply a Gradient Reversal Layer (GRL)~\cite{qu2025disentanglement} to $a_{c}$, \ie $\tilde{a} = \operatorname{GRL}(a_c,{\alpha}_{c})$, where $\alpha_{c}$ is a hyperparameter controlling the strength of the gradient reversal, and then feed $\tilde{a}$ to the style classifier $C(\cdot)$ to obtain predicted logits $\hat{y} = C(\tilde{a})$.
The adversarial loss $\mathcal{L}_{\text{adv}}$ is computed as the cross-entropy loss between the predicted labels $\hat{y}$ and the true style labels $y$:
\begin{equation}
    \mathcal{L}_{\text{adv}} = \mathrm{CE}(\hat{y}, y).
\end{equation}

\textbf{Style Similarity Loss.} Given the style embeddings $s_a$, $s_m$, and $s_p$, the style similarity loss is calculated based on cosine similarity. The cosine similarity between $s_a$ and $s_m$ is defined as:
\begin{equation}
    \text{cosine}(s_a, s_m) = \frac{s_a \cdot s_m}{\Vert s_a\Vert \Vert s_m \Vert},
\end{equation}
where $s_a \cdot s_m$ is the dot product, and $\Vert \cdot \Vert$ denotes the Euclidean norm. The cosine similarity loss aims to maximize the similarity, and $\mathcal{L}_{\text{cos}}$ is defined as:
\begin{equation}
\begin{aligned}
    \mathcal{L}_{\text{cos}} =  w_1 \left(1 - \text{cos}(s_a,s_m)\right) & + 
   w_2 \left(1 - \text{cos}(s_a, s_p)\right) +\\
   w_3 \left(1 - \text{cos}(s_m, s_p)\right),
\end{aligned}
\end{equation}
where $w_1$, $w_2$, and $w_3$ are weights of the style embeddings.

\textbf{Orthogonality Loss and Mutual Information Minimization Loss.} Given the content features $C \in \mathbb{R}^{B \times F_c \times D_c}$  and style features $S \in \mathbb{R}^{B \times F_s \times D_s}$ extracted from motion sequences or input audio, the goal is to align the frame counts of the two feature sets and minimize their inner product to enforce orthogonality. To match the frame dimensions between content and style features, we apply adaptive average pooling to ensure consistency in the number of frames across both feature matrices. After aligning the frames, we compute the inner product between content and style features. Let $\textbf{\textit{P}} \in \mathbb{R}^{B \times D_c \times D_s}$ denote the batch-wise inner product, where $B$ is the batch size, $F_c$ and $F_s$ are the number of frames, and $D_c$ and $D_s$ are the feature dimensions. Then, we apply a modified Procrustes loss to encourage orthogonality, namely:
\begin{equation}
    \mathcal{L}_{\text{orth}} = \frac{1}{B} {\Vert \textbf{\textit{P}} \textbf{\textit{P}}^\top - \textbf{\textit{I}} \Vert}_{\text{Frob}},
\end{equation}
where $\textbf{\textit{I}}$ is an identity matrix, Frob is the Frobenius norm, and  ${\Vert \textbf{\textit{P}} \textbf{\textit{P}}^\top - \textbf{\textit{I}} \Vert}_{\text{Frob}}$ enforces disentanglement by minimizing the inner product between the content and style features. 

Furthermore, orthogonality constraints encourage the feature vectors to be at right angles, and they only regulate the angular relationship and do not directly reduce the amount of shared information between the modalities. Therefore, we leverage a mutual information minimization loss. Instead of simply constraining the vector angles, the mutual information minimization loss is designed to explicitly reduce the statistical dependency between the content representation and the style representation. The mutual information minimization loss $\mathcal{L}_{\mathrm{Info}}$ is then formulated as:
\begin{equation}
    \mathcal{L}_{\mathrm{Info}} = -\frac{1}{B}\sum_{i=1}^{B} \log \frac{\exp\left(\frac{\mathrm{sim}(a_i, c_i)}{\tau}\right)}
    {\sum_{j=1}^{B} \exp\left(\frac{\mathrm{sim}(a_i, c_j)}{\tau}\right)},
\end{equation}
where $\tau$ is a temperature hyperparameter. 

\textbf{Top-$\bm{k}$ Bidirectional Contrastive and KL divergence Loss.}  We adopt a Top-$k$ bidirectional contrastive loss and a KL divergence loss to align the content features of audio and motion sequences at the feature level and reduce the modality gap. Given motion content encoding $C \in \mathbb{R}^{B \times F_c \times D_c}$ and audio feature $A \in \mathbb{R}^{B \times F_a \times D_a}$, we compute the similarity matrices for both audio-to-content $E^{A \to C}$ and content-to-audio $E^{C \to A}$. 

For each audio query vector $A_i$, we define that the positive sample is the content encoding $C_i$ and the negative samples are other $C_j$ (for $i \neq j$). To address the imbalance between positive and negative samples from different modalities (audio and mesh), we introduce a weighting mechanism to adjust their contributions in contrastive loss. Specifically, we use weighting factors $\alpha$ for positive samples and  $\beta$ for negative samples. This balances their contributions in contrastive loss. Then, we use a Top-$k$ strategy in contrastive loss, considering only the top $k$ most similar key vectors for each query vector. We calculate the contrastive loss for both audio-to-content $\mathcal{L}_{a \to c}$ and content-to-audio $\mathcal{L}_{c \to a}$:
\begin{small}
\begin{equation}
\begin{aligned}
\mathcal{L}_{a \to c} &= \frac{1}{B} \sum_{i=1}^{B} \left( -\alpha \cdot \log \frac{\exp(E^{A \to C}_{i,i})}{\beta \cdot \sum_{j \in \text{Top-}k(E^{A \to C}_i)} \exp(E^{A \to C}_{i,j})} \right),\\ 
\mathcal{L}_{c \to a} &= \frac{1}{B} \sum_{i=1}^{B} \left( -\alpha \cdot \log \frac{\exp(E^{C \to A}_{i,i})}{\beta \cdot \sum_{j \in \text{Top-}k(E^{C \to A}_i)} \exp(E^{C \to A}_{i,j})} \right).
\end{aligned}
\end{equation}
\end{small}
The total contrastive loss is the weighted sum of the audio-to-content and content-to-audio losses:
\begin{equation}
\mathcal{L}_{\text{Top-$\bm{k}$}} = \lambda \mathcal{L}_{a \to c} + (1 - \lambda) \mathcal{L}_{c \to a},
\end{equation}
where $\lambda$ is a coefficient that balances the contributions.

$\mathcal{L}_{\text{Top-$\bm{k}$}}$ is effective at enforcing local similarity by selecting the most similar (or ``hard'') negatives for each sample, but this local focus does not guarantee that the overall statistical distributions of the features across different modalities are consistent. Therefore, we further utilize the KL divergence loss that aligns the global distributions of the content features extracted from the audio modality and the motion features extracted from the facial movements. The representations of audio and motion are $a$ and $c$, and the KL divergence from the audio to the motion is given by:
\begin{equation}
    \mathcal{L}_{\mathrm{KL}} = \frac{1}{2}\sum_{d=1}^{D}\left[ \log\frac{\sigma_{c,d}^2}{\sigma_{a,d}^2} + \frac{\sigma_{a,d}^2 + (\mu_{a,d}-\mu_{c,d})^2}{\sigma_{c,d}^2} - 1 \right].
\end{equation}

Minimizing $\mathcal{L}_{\mathrm{KL}}$ encourages the overall distributions to be aligned. This global alignment complements the local constraints imposed by the Top-$k$ contrastive loss, leading to a more robust cross-modal feature alignment. Therefore, the final contrastive loss $\mathcal{L}_{\text{cts}}$ is:
\begin{equation}
    \mathcal{L}_{\text{cts}} = \mathcal{L}_{\text{Top-$\bm{k}$}} + \mathcal{L}_{\text{KL}}.
\end{equation}

\textbf{Reconstruction Loss and Mouth Loss.} $\mathcal{L}_{\text{rec}}$ and $\mathcal{L}_{\text{mou}}$ represent the loss for reconstructing the sequences and the mouth region, 
\ie:
 \begin{equation}
  \mathcal{L}_{\text{rec}} = \frac{1}{N} \sum_{n=1}^{N} \frac{1}{T_{n}} \sum_{t=1}^{T_{n}} \Arrowvert d_{n,t}-\hat{d}_{n,t} \Arrowvert_{2}, 
 \end{equation}
  \begin{equation}
  \mathcal{L}_{\text{mou}} = \frac{1}{N} \sum_{n=1}^{N} \frac{1}{T_{n}} \sum_{t=1}^{T_{n}} \Arrowvert m_{n,t}-\hat{m}_{n,t} \Arrowvert_{2},
 \end{equation}
where $N$ is the number of sequences, $T_{n}$ denotes the length of the $n$-th sequence, and ($d$, $m$) and ($\hat{m}$, $\hat{d}$) represent ground truth and the corresponding prediction, respectively. 

\textbf{Temporal Consistency Loss.} The velocity loss $\mathcal{L}_{vel}$ aims to increase temporal consistency, \ie:
 \begin{equation}
  \mathcal{L}_{\text{vel}} = \frac{1}{N} \sum_{n=1}^{N} \frac{1}{T_{n}} \sum_{t=2}^{T_{n}} \Arrowvert (d_{n,t} - d_{n,t-1}) - (\hat{d}_{n,t}-\hat{d}_{n,t-1})\Arrowvert_{2}.
 \end{equation}
Therefore, the whole motion loss $\mathcal{L}_{\text{motion}}$ is:
 \begin{equation}
  \mathcal{L}_{\text{motion}} = \alpha_{1}\mathcal{L}_{\text{rec}}+\alpha_{2}\mathcal{L}_{\text{mou}}+\alpha_{3}\mathcal{L}_{\text{vel}},
 \end{equation}
where $\alpha_{i}$ controls the contributions of each loss term.

\textbf{Training Objects.} Combining the well-designed constraints with the motion loss, the whole loss can be calculated by:
 \begin{equation}
  \mathcal{L}_{\text{all}} = \beta_{1}\mathcal{L}_{\text{motion}}+\beta_{2}\mathcal{L}_{\text{adv}}+\beta_{3}\mathcal{L}_{\text{orth}}+\beta_{4}\mathcal{L}_{\text{Info}}+\beta_{5}\mathcal{L}_{\text{cts}},
 \end{equation}
where the hyper-parameters $\beta_{1}$, $\beta_{2}$, $\beta_{3}$, $\beta_{4}$, and $\beta_{5}$ control the contribution degree of each loss term.

\subsection{Modality Alignments}
The fundamental differences between audio and 3D mesh data, \ie, where audio is a time-domain signal and mesh data represents spatial geometric structures, raise alignment challenges due to their diverse nature in feature space.

To address spatial misalignment, we treat the 3D mesh as a graph structure, using a graph attention encoder $E_{G}$ to extract topological features from the mesh. The $E_{G}$ process can be defined as:
\begin{equation}
    h_v^{(l+1)} = \sigma\left( \sum_{u \in \mathcal{N}(v)} \alpha_{vu} \cdot W^{(l)} h_u^{(l)} \right),
\end{equation}
where $h_v^{(l)}$ is the feature of node $v$ at layer $l$, $\mathcal{N}(v)$ represents the neighbors of node $v$, $\alpha_{vu}$ is the attention weight between nodes $v$ and $u$, and $W^{(l)}$ is the learnable weight matrix at layer $l$. The output features are then fused with audio features in a shared spatial embedding space. This fusion allows spatially grounded information transfer between the modalities.
For temporal alignment, we employ a Transformer with cross-modal attention mechanisms $\mathcal{D}$ to capture temporal dependencies between audio and mesh data, enabling the model to learn complex temporal relationships and achieve sequence synchronization. The process is formulated as:
\begin{equation}
   \hat{M}_{1:T} = \mathcal{D}(E_a^c(\chi), E_m^c(E_{G}(M_{1:T})), s).
\end{equation}
Additionally, by employing Top-$k$ bidirectional contrastive learning loss and the KL loss, \ie $L_{\text{cts}}$, we map both modalities into a content space and optimize the embeddings to reduce the modality gaps, namely $\mathcal{L}_{\text{cts}}(a_{1:T},c_{1:T})$. This three-level alignment strategy (\ie, addressing spatial, temporal, and feature space differences) enhances the integration of audio and mesh embeddings, thereby improving lip-sync accuracy for 3D talking head animation.

\section{Experiments}
\label{sec:exp}

\subsection{Datasets and Implementations}
\textbf{Datasets.} We conducted experiments using two publicly available datasets, VOCASET and BIWI,  including audio-3D scan pairs with English speech pronunciation. \textbf{VOCASET Dataset} includes 480 paired audio-visual sequences recorded from 12 subjects, including six males and six females. The facial motion is captured at 60\,fps, with each 3D face mesh represented by 5\,023 vertices with a registered topology. \textbf{BIWI Dataset} is a 3D audio-visual corpus of affective speech and facial expression comprising dense dynamic 3D face geometries. Fourteen subjects, including six males and eight females, were asked to say 40 English sentences twice with emotion or not. The 3D face dynamics are captured at 25\,fps, and the registered topology contains 23\,370 vertices. Specifically, BIWI contains two test sets: BIWI-Test-A includes 24 sentences spoken by six seen subjects for quantitative and qualitative evaluation; BIWI-Test-B contains 32 sentences spoken by eight unseen subjects, which is more suitable for qualitative evaluation. For a fair comparison, we follow the same settings in~\cite{fan2022faceformer,xing2023codetalker,song2024talkingstyle} for both the VOCASET dataset and the BIWI dataset.

\textbf{Implementations.} 
We sampled the audio at the rate of 16\,kHz. We trained the network on an NVIDIA RTX 3090 and employed the Adam optimizer for updating the model parameters with an initial learning rate of $3 \times 10^{-4}$. We employed state-of-the-art methods for comparison with our approach. 

\textbf{Training Strategy.} For a more complete disentanglement, our training process gradually unfroze the style decoders. Specifically, in the first training stage, the style encoder is kept frozen so that only the content encoder and the style classifier are trained adversarially. This forces the content encoder to generate representations that do not carry discriminative style information. In the second stage, we unfreeze the style encoder so that it can learn detailed style information, while the content encoder is still encouraged (via GRL) to remove style cues. This two-stage training strategy helps achieve a better separation between content and style. 

\begin{table}[!t]
\begin{center}
\caption{Quantitative evaluation results on the {VOCASET-Test}. The results with the superscript * are reproduced by us. 
}
\label{tab:vocaset}
\begin{threeparttable}
\setlength{\tabcolsep}{4pt}
\begin{tabular}{cccc}
\toprule
\multirow{2}{*}{Methods} & \multirow{2}{*}{\begin{tabular}[c]{@{}c@{}}LVE$\downarrow$\\ ($\times 10^{-5}$mm)\end{tabular}} & 
\multirow{2}{*}{\begin{tabular}[c]{@{}c@{}}FDD$\downarrow$\\($\times 10^{-7}$mm)\end{tabular}} & \multirow{2}{*}{LRP$\uparrow$} \\&          &          &         \\ 
\midrule
VOCA~\cite{cudeiro2019capture}       & 4.9245   & 4.8447   & 72.67\,\%    \\
MeshTalk~\cite{richard2021meshtalk}  & 4.5441   & 5.2062   & 79.64\,\%    \\
FaceFormer~\cite{fan2022faceformer}  & 4.1090   & 4.6675   & 88.90\,\%    \\
CodeTalker~\cite{xing2023codetalker} & 3.9445   & 4.5422   & 86.30\,\%    \\
FaceDiffuser~\cite{stan2023facediffuser} & 3.2432   & 4.4326   & -    \\
DiffSpeaker~\cite{ma2024diffspeaker} & \underline{3.1478}   & -   &  -    \\
SelfTalk~\cite{peng2023selftalk}     & 3.2238   & \underline{4.0912}  & \underline{91.37}\,\%     \\
TalkingStyle~\cite{song2024talkingstyle} & 3.3159*   & 4.2874*  & -           \\ 
NewTalker~\cite{niu2025exploring} & 3.2026 & - & -           \\
\midrule
\textbf{PTalker (Ours)}   & \textbf{2.8856}  & \textbf{3.7325}  & \textbf{93.77\,\% }   \\ 
\bottomrule
\end{tabular}
\end{threeparttable}
\end{center}
\end{table}

\begin{table}[!t]
\begin{center}
\caption{Quantitative evaluation results on the BIWI-Test-A. 
}
\label{tab:biwi}
\begin{threeparttable}
\setlength{\tabcolsep}{4pt}
\begin{tabular}{cccc}
\toprule
\multirow{2}{*}{Methods} & \multirow{2}{*}{\begin{tabular}[c]{@{}c@{}}LVE$\downarrow$\\ ($\times 10^{-4}$mm)\end{tabular}} & 
\multirow{2}{*}{\begin{tabular}[c]{@{}c@{}}FDD$\downarrow$\\($\times 10^{-5}$mm)\end{tabular}} & \multirow{2}{*}{LRP$\uparrow$} \\  &          &          &     \\ 
\midrule
VOCA~\cite{cudeiro2019capture}       & 6.5563    & 8.1816   & 73.83\,\%    \\
MeshTalk~\cite{richard2021meshtalk}  & 5.9181   & 5.1025   & 80.97\,\%    \\
FaceFormer~\cite{fan2022faceformer}  & 5.3077   & 4.6408   & 83.15\,\%    \\
CodeTalker~\cite{xing2023codetalker} & 4.7914   & 4.1170   & 84.62\,\%    \\
FaceDiffuser~\cite{stan2023facediffuser} & 4.7823   & 3.9225   & -    \\
DiffSpeaker~\cite{ma2024diffspeaker} & 4.2829   & 3.6823   &  -    \\
SelfTalk~\cite{peng2023selftalk}     & 4.2485  & 3.5761 & \underline{88.31}\,\%     \\
TalkingStyle~\cite{song2024talkingstyle} & 4.3646   & 3.8361  & -           \\ 
NewTalker~\cite{niu2025exploring} & 4.2361 & 3.4903  & -           \\
DCPTalk~\cite{chu2025speech} & \underline{3.8548} & \underline{2.8067}  & -           \\
\midrule
\textbf{PTalker (Ours)}   & \textbf{3.8211}  & \textbf{2.5493}  & \textbf{91.68\,\% }   \\ 
\bottomrule
\end{tabular}
\end{threeparttable}
\end{center}
\end{table}

\begin{figure*}[!ht]
  \centering
   \includegraphics[scale=0.52]{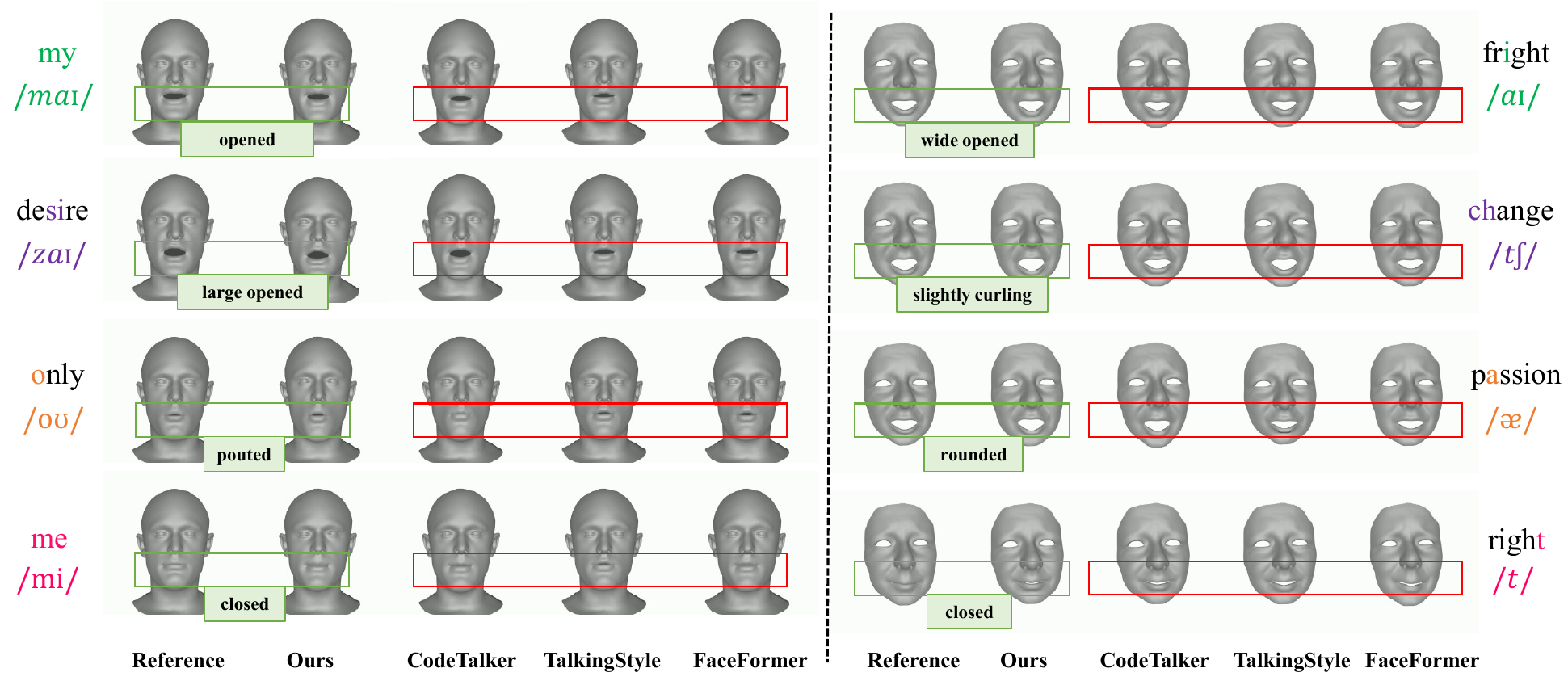}
   \caption{Visualization results on the VOCASET-Test (left) and the BIWI-Test-B (right). We compare the mouth movements on some key syllables generated by our method against competitors. 
   }
   \label{fig:quality}
   \Description{.}
\end{figure*}

\subsection{Quantitative Evaluation}
\textbf{Metrics} Following~\cite{fan2022faceformer,xing2023codetalker}, we use Lip Vertex Error (LVE) and Upper-face Dynamics Deviation (FDD) to measure lip-sync and overall facial expression, respectively. Additionally, we use lip readability percentage (LRP)~\cite{peng2023selftalk} to reflect the lip readability. Based on the results in Tab.~\ref{tab:vocaset} and Tab.~\ref{tab:biwi}, our PTalker method consistently outperforms state-of-the-art methods on both VOCASET-Test and BIWI-Test-A datasets, achieving lower errors and LRP performance, demonstrating the best lip-sync accuracy. Specifically, PTalker demonstrates a reduction in lip vertex error and face dynamic distance metrics, reflecting improved precision in capturing facial and lip movement details, namely lip-sync and speaking style. Notably, PTalker improves FDD by 10.10$\,\%$ over the next best method, DCPTalk, on the BIWI-Test-A dataset, demonstrating the effectiveness of our style disentanglement and three-level alignment strategy.

\subsection{Qualitative Evaluation}
We visually compare our PTalker with other methods. Fig.~\ref{fig:quality} highlights the effectiveness of our method in generating 3D talking head animations with precise lip-sync and style capture. Our PTalker synchronizes lip movements with phonemes more accurately than other methods. For example, it captures the dynamics of mouth opening for `/a\i/' in ``my'' and ``fright'', and mouth closure for `/mi/' for ``me'' and `/t/' in ``right''. Additionally, our PTalker adeptly replicates the nuances of speaking styles, including rounded (\textipa{`/ae/'} in ``passion''), pouted (\textipa{`/oU/'} in ``only''), and slightly curling (\textipa{`/tS/'} in ``change'') lip shapes. These capabilities empower PTalker to generate a more lifelike and expressive animation, showcasing the superior performance of our method in capturing the subtleties of individual speaking styles.

\subsection{Speaking Style Visualization}
\begin{figure}[t]
  \centering
   \includegraphics[scale=0.2]{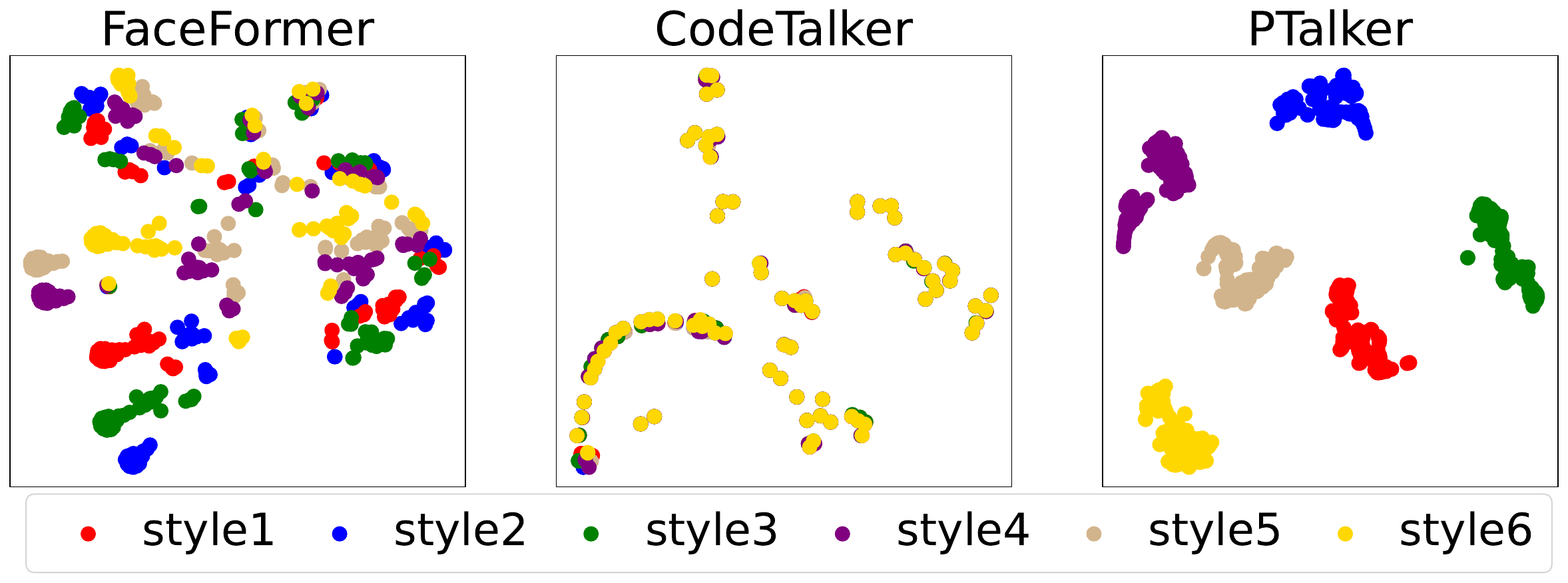}
   \caption{Visual comparison results under different personal styles with the same sentence by the t-SNE on BIWI-Test-B. 
   }
   \label{fig:style}
   \Description{.}
\end{figure}

\begin{figure*}[!t]
\centering
  \includegraphics[scale=0.51]{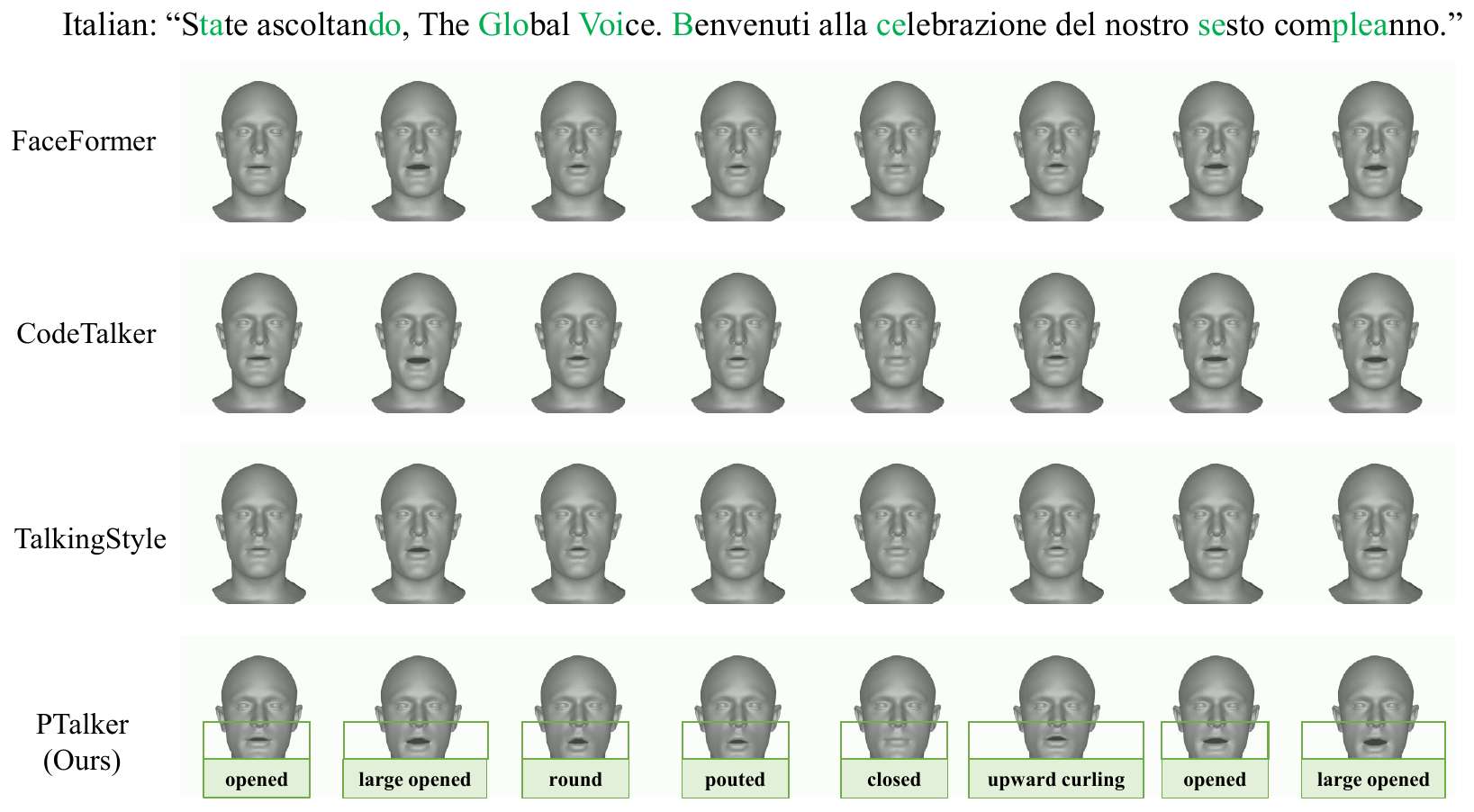}
  \caption{Visual comparisons of the generalization evaluation using an Italian speech on the VOCASET dataset.}
  \label{fig:ital}
  \Description{.}
\end{figure*}

In the t-SNE~\cite{van2008visualizing} visualization, each point represents a frame, with colors denoting different speaker styles. Fig.~\ref{fig:style} reveals how each method performs in clustering speaking styles. FaceFormer (left panel in Fig.~\ref{fig:style}) shows some distinction among styles but struggles with clear separation. These overlap points can be attributed to fixed identity binding, which limits its generalization to unseen subjects. CodeTalker (middle panel in Fig.~\ref{fig:style}) unbinds identity from speech, enhancing the facial performance of unseen ones. However, this approach compromises the retention of individual speaker styles, as the styles converge in the same position. PTalker (right panel in Fig.~\ref{fig:style}) excels at maintaining clear clusters for each style, with distinct color separations. This indicates that PTalker effectively captures and preserves each speaker’s unique characteristics and generalizes well to unseen ones.

\subsection{Generalization Evaluation}

To evaluate PTalker's generalization capability, we tested it with samples outside the VOCASET and BIWI datasets. For example, we used an Italian speech to generate 3D talking head animations. The visual results are shown in Fig.~\ref{fig:ital}. As observed, PTalker more accurately and naturally simulates the shape changes of the lips during speech than competitors. In particular, PTalker demonstrates superior performance in the following aspects: more pronounced upper lip curling and pouted mouth in the case of ``voice'' and ``celebrazione'', tighter lip closures in ``Benvenuti'', and larger mouth openings in ``compleanno''. These observations indicate that PTalker generates more natural lip dynamics, effectively capturing subtle lip movements across diverse speech inputs.

\subsection{User Study}
\begin{table}[t]
\setlength{\tabcolsep}{3pt}
\caption{User study results on the VOCASET-Test and BIWI-Test-B datasets. Higher values are better for both metrics.}
\label{tab:user}
\centering
\begin{threeparttable}
\begin{tabular}{lcccc}
\hline
\multirow{2}{*}{\textbf{Competitors}} & \multicolumn{2}{c}{\textbf{VOCASET-Test}} & \multicolumn{2}{c}{\textbf{BIWI-Test-B}} \\
\cline{2-5}
 & \textbf{Lip Sync}$\uparrow$ & \textbf{Realism}$\uparrow$ & \textbf{Lip Sync}$\uparrow$ & \textbf{Realism}$\uparrow$ \\
\hline
FaceFormer & 3.27 &  3.12 &  3.41 &  3.62 \\
CodeTalker & 3.42 & 3.56  & 3.68  & 3.74  \\
TalkingStyle &  3.39 &   3.42 & 3.74 & 3.86 \\
\textbf{Ours} &  \textbf{3.78}  &   \textbf{4.05} & \textbf{4.08} & \textbf{3.92} \\
\hline
\end{tabular}
\end{threeparttable}
\end{table}

\begin{table}[t]
\centering
\caption{Ablation studies on the BIWI-Test-A dataset.}
\label{tab:ab2}
\begin{threeparttable}
\setlength{\tabcolsep}{5pt}
\begin{tabular}{llll}
\hline
     & LVE$\downarrow$ & FDD$\downarrow$ & LRP$\uparrow$ \\ \hline
\textbf{Ours} & \textbf{3.8211}  & \textbf{2.5493} &  \textbf{91.68\,\% }    \\ 
w/o $\mathcal{L}_{\text{adv}}$  &  4.0093   &  2.8528     &    90.87\,\%  \\
w/o $\mathcal{L}_{\text{cos}}$  & 3.9821 &    2.6309    &   91.02\,\%  \\
w/o $\mathcal{L}_{\text{orth}} + \mathcal{L}_{\text{Info}}$ &  4.1792   &    3.2826     &    89.45\,\%  \\
w/o $\mathcal{L}_{\text{cts}}$  &   4.3107    &  3.3024   &    88.64\,\%   \\
w/o $\mathcal{L}_{E_{G}}$   &   4.0612   & 2.8951  &    90.85\,\%  \\
w/o Audio Disentanglement  &  4.2356  &   3.2157  &    89.13\,\%  \\
w/o Motion Disentanglement & 4.2844  &  3.2679   &   88.92\,\%   \\ \hline
\end{tabular}
\end{threeparttable}
\end{table}

To provide a more comprehensive evaluation of PTalker, we conducted user studies, similar to those in ~\cite{peng2023selftalk, song2024talkingstyle}. 48 participants were asked to rate the talking head animation videos among PTalker and competitors, on a scale of 1-5, with higher scores indicating a better preference. Specifically,  the 48 participants consist of 28 males and 20 females, with the primary age group being 18 to 30 years (33 individuals), and the remaining 15 users aged between 35 and 50. We designed a total of 9 comparisons, with three in each group, and we calculated the Mean Opinion Score (MOS). Tab.~\ref{tab:user} shows that more users favor PTalker, suggesting that the animation videos of PTalker have superior visual quality.

\subsection{Ablation Study}
We conducted ablation studies to evaluate the impact of removing various components from our model, as shown in Tab.~\ref{tab:ab2}. The results demonstrate that the full model configuration consistently achieves the best performance on the BIWI-Test-A, emphasizing the importance of each component. Specifically, the absence of $\mathcal{L}_{\text{adv}}$, $\mathcal{L}_{\text{cos}}$, $\mathcal{L}_{\text{orth}}$, $\mathcal{L}_{\text{Info}}$, and $\mathcal{L}_{\text{cts}}$ leads to increases in LVE and FDD, which suggests that these losses are critical for reducing the modality gap and effectively disentangling the style and content spaces. Moreover, removing $E_G$ also results in higher LVE and FDD values, confirming the importance of spatial alignment introduced by $E_G$. Removing audio or motion disentanglement degrades the overall performance, as evidenced by increased LVE and FDD and decreased LRP. These findings highlight the necessity of disentanglement for achieving high-quality and expressive animations.

\section{Conclusion and Future Works}
\label{con}

In this work, we present PTalker, a framework for generating personalized speech-driven 3D talking head animations by disentangling style and aligning modalities. With well-designed constraints, PTalker ensures that the generated animations are accurately synchronized with speech and effectively match the corresponding speaking style. Extensive qualitative and quantitative experiments demonstrate the effectiveness of our approach. We believe that PTalker will contribute to the development and application of personalized 3D talking head animations. However, due to limitations in high-quality 3D facial data, current disentanglement algorithms face difficulties in fully separating style codes and content space. A promising direction for future work is addressing the issue of incomplete disentanglement.

\begin{acks}
The work leading to this research was supported by the Guangdong Basic and Applied Basic Research Foundation under Grant No.~2024A1515010112 and the Changsha Science and Technology Bureau Foundation under Grant No.~kq2402082. It was also partically supported by the National Natural Science Foundation of China under Grant No.~62076092, 62272154, 62522208, and 62472163, the Hunan Provincial Natural Science Foundation of China under Grant No.~2024JJ5096, the science and technology innovation Program of Hunan Province under Grant No.~2023RC3125, the Science \& Technology talents lifting project of Hunan Province under Grant No.~2023TJ‒N23, the Training Program for Excellent Young Innovators of Changsha under Grant No.~kq2209008, and the Hunan Provincial Department of Education Excellent Youth Project under Grant No.~24B0039.  
\end{acks}

\balance
\bibliographystyle{ACM-Reference-Format}
\bibliography{sample-base}

\end{document}